# Dynamic Hierarchical Bayesian Network for Arabic Handwritten Word Recognition

Khaoula jayech [#1], Nesrine Trimech [#2], Mohamed Ali Mahjoub [#3], Najoua Essoukri Ben Amara [#4]

[#] Research Unit SAGE , National School of Engineers of Sousse, University of Sousse, BP 264, Sousse Erriadh, 4023, Tunisia

[1] jayech_k@yahoo.fr
[2] nesrine.trimech@gmail.com
[3] medali.mahjoub@ipeim.rnu.tn
[4] najoua.benamara@eniso.rnu.tn

*Abstract*— **This paper presents a new probabilistic graphical model used to model and recognize words representing the names of Tunisian cities. In fact, this work is based on a dynamic hierarchical Bayesian network. The aim is to find the best model of Arabic handwriting to reduce the complexity of the recognition process by permitting the partial recognition. Actually, we propose a segmentation of the word based on smoothing the vertical histogram projection using different width values to reduce the error of segmentation. Then, we extract the characteristics of each cell using the Zernike and HU moments, which are invariant to rotation, translation and scaling. Our approach is tested using the IFN / ENIT database, and the experiment results are very promising.**

I. INTRODUCTION

Offline Arabic handwriting recognition (HWR) denotes the process of transforming the Arabic text, which is presented in its spatial form of graphical marks, into its symbolic representation [10]. The recognition of Arabic handwritten words has many applications in bank check reading, mail sorting, and form processing in administration and insurance [11],[12],[13]. The major problem of Arabic handwriting recognition can be divided into two important research areas: segmentation and recognition. These two areas are difficult tasks due to the high variability and uncertainty of human writing (shape discrimination and variation, scanning methods, fusion of diacritical points, writing discontinuity and slant, overlapping, touching of sub-words…), which makes the process of word Arabic segmentation and recognition very difficult.

In this paper, we present a methodology that segments a word and recognizes it. The recognition algorithm is preceded by a feature extraction algorithm that extracts the character and sub-character features from the segmented word. To segment the Arabic word, we use a smoothed vertical histogram projection with different width values to minimize the error of segmentation.

The rest of the paper is divided as follows. In section 2, we describe works related to our research. In section 3, we describe the architecture of our system and the concepts of dynamic hierarchical Bayesian networks. Finally, we present the obtained results and conclusions.

II. REVIEW OF OFFLINE HANDWRITTEN WORD RECOGNITION APPROACHES

A review of the literature shows that relatively limited applications, based on Bayesian Networks (BNs) and Dynamic Bayesian Networks (DBNs), have been developed for the handwriting recognition. In what follows, we state the main works.

Hallouli in [1] [2] developed a new probabilistic model designed for off-line printed character recognition based on the DBN. Their model consisted in coupling two Hidden Markov Models (HMMs) in various DBN architectures. The first HMM model was obtained from observations of pixel-image columns (vertical-HMM), and the second from observations of image rows (horizontal-HMM). Their system was evaluated using various DBN architectures and achieved a recognition rate of 98.3% with the vertical HMM, and 93.7% with the horizontal one. However, when testing degraded letters the recognition rate went down to 93.8% with the vertical HMM and 88.1% with the horizontal one. This modeling allows overcoming the HMM limitations while getting a good modeling of handwritten character images, due to the combinations of information about the rows and columns. Nevertheless, the choice of coupling links should be well studied by learning a structure which is a difficult task in the DBN.

Similarly, Mahjoub et al. in [4] proposed a new system for the offline handwritten Arabic word recognition based on coupled HMMs, considered as a single DBN. Each image of the handwritten word was transformed into two sequences of feature vectors that would be the observations to be given to the DBN model; the first vector feature sequence modeled the flow of observations on the columns and the second one modeled the flow of observation on the rows. The developed system was tested on the IFN/ENIT database and achieved a recognition rate of 67.9% -77.4%.

AlKhateeb in [13] proposed an offline recognition system based on HMMs. The method was composed of three stages: preprocessing, feature extraction and classification. First, words were segmented and normalized. Then, a set of intensity features were extracted from each of the segmented words using a sliding window moving across each mirrored word image. Finally, a combined system was developed using these features for classification. Intensity features are used to train an HMM classifier, whose results are re-ranked using structure-like features to improve the recognition rate. To validate the proposed techniques, extensive experiments were carried out using the IFN/ENIT database. The results were very promising.

El-Hajj in [11] presented a system for the off-line recognition of handwritten Arabic city names. The system was based on the HMM applied with a set of features including both the independent baseline and the dependent one. The features were extracted using the sliding window. The system combined three homogeneous HMMs having the same topology as the reference system and differ only in the orientation of the sliding window. The results showed that the combination of classifiers would perform better than a single classifier dealing with slant-corrected images, and that the approach was robust for a wide range of orientation angles.

Parvez in [14] proposed an off-line Arabic handwritten text recognition system using structural techniques. An Arabic text line was segmented into words/sub-words and dots were extracted leaving the Parts of Arabic words (PAW). Afterwards, the PAW were slant-corrected using a novel slant estimation and correction technique based on a polygonal approximation that was robust and had a high resolution. Then, the segmentation algorithm was integrated into the recognition phase of the handwritten text. After that, the Arabic characters were modeled by 'fuzzy' polygons. The recognition of the Arabic PAWs was done using a novel fuzzy polygon matching algorithm. This proposed system was tested using the IFN/ENIT database, and the recognition rate achieved was promising, which was about 79.58%.

The analysis of the literature shows that many works applied the HMMs to model and recognize the offline handwritten Arabic word and limited research has discovered the DBNs. So, we have proposed in Jayech (2012) [3] a new system based on the hierarchical BN as a global approach to recognize the word image without any segmentation. The aim of this approach is to find the best model of Arabic handwriting to reduce the complexity of the recognition process by permitting the partial recognition . As first experiments, we have obtained promising results. As a continuation to our research, we have proposed to segment the Arabic word into characters and sub-characters using the smoothed vertical histogram projection, and then, search the best approach that can model the temporal aspect. In this way, we propose to use the Dynamic Hierarchical Bayesian Networks (DHBNs) which make good results in the human interaction domain, but it was never been used in the Arabic handwriting recognition.

The contributions of this paper are as follows: (1) A new hierarchical framework is proposed for the recognition of the off-line Arabic handwritten word at a detailed level using the HBN. (2) A stochastic graphical model is proposed for the word recognition. (3) A combined system that balances between segmentation and recognition is proposed to fit and decrease the error of segmentation to have a higher recognition rate.

## III. SYSTEM ARCHITECTURE

In this paper, we have developed an off-line recognition system for the handwritten Arabic city names using DHBNs. The system architecture contains five stages in terms of preprocessing, segmentation, feature extraction and vector quantization, learning and classification in the following sections. The proposed block diagram is illustrated in Fig.1.

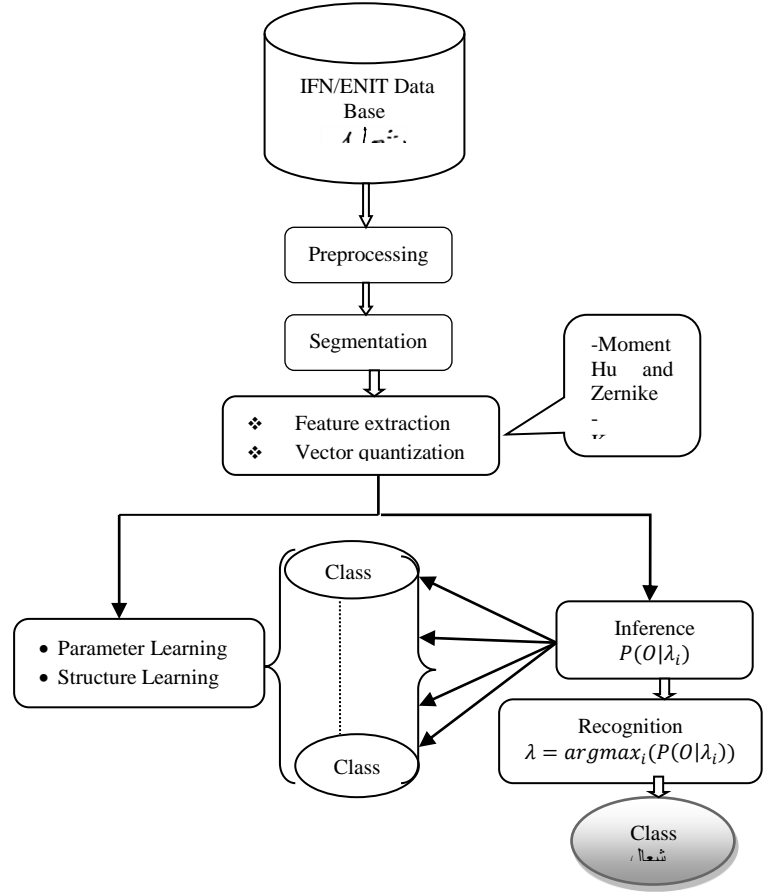

Fig. 1 System architecture showing the main stages which must be carried out to identify the word image.

As shown in Fig.1, The input image goes through the preprocessing steps to standardize the signal. Afterwards, the word image of the city name has been segmented, using the sliding window approach based on the vertical projection histogram analysis of the word, into characters that are then divided into a set of frames and cells. After that, the features are extracted, quantized and fed to a DHBN classifier in order to classify an unknown word by deciding to which class it belongs.

### A. Preprocessing and Segmentation

After scanning a document, some basic preprocessing tasks like image normalization, word segmentation and noise reduction have to be performed. Due to the fact that we use the crapped binary word images coming from the IFN/ENIT database, some of these fundamental preprocessing tasks have already been done during the database development. The processing chain of the recognizer starts out with some of the usual operation such as: remove space and remove diacritical marks. Subsequently, in the second step, we segment the word image. Consciously, the segmentation is a difficult task in the Arabic handwritten word recognition, due to its high variability, especially when dealing with a large lexicon for the semi cursive scripts as Arabic. We segment the Arabic handwritten word into characters and sub-characters. The general idea is to find the best segmentation of the word image, without any diacritical marks, into characters using the smoothed vertical projection histogram. The pre-processing and segmentation algorithm is presented in Fig.2 and Fig.3:

---

Pre-Processing and segmentation algorithm

For k=1 to number of images of each word

  Img_in = Read the binary image.
  Img_in = Resize the Img_in into $100 \times 200$.
  Img_in = Remove the horizontal and vertical space from image.
  Img_in_Diac = Remove the diacritical marks from Img_in.
  Blocks_Limit = Segment the image word into characters by finding the boundaries of each character using the peaks of the smoothed vertical projection histogram of Img_in_Diac finding by a fixed optimal value of width of smoothing.
  Bloc_Charact = Divide the Img_in using each Blocks_Limit into characters then divide each characters into 3 uniform horizontal frame and each frame into 2 uniform cell.
END

---

Fig. 2   Pre-processing and segmentation algorithm.

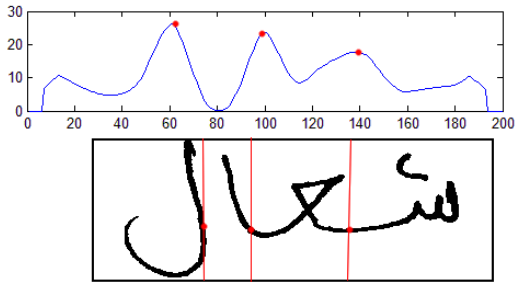

Fig. 3   Word segmentation into characters: (a) Smoothed vertical projection histogram, (b) Blocks of characters

### B. Feature extraction and vector quantization

The feature extraction process starts by splitting the character image into a set of cells, using the sliding window, with a fixed width in a horizontal direction to create 3 frames divided into 2 cells. The choice of the optimal number of the cell is determined empirically yielding the best recognition results. For each cell we extract the moment invariants of the Zernike and HU descriptors, which are invariant to rotation, translation and scaling [7], to check the characteristics of each character. The descriptors of the moment invariants give us continuous signatures. However, the DBNs require discrete variables. So, we process to the next stage of pre-treatment, which consist discreting the obtained variables using the k_means method [9].

### C. Classification

The DBNs are a class of temporal graphical probabilistic models that have become a standard tool for modeling various stochastic time varying phenomena. The temporal probabilistic graphical models as two-time BN are the most used and popular models for the DBN. Before introducing the notion of the DBN, we will briefly recall the definition of BN.

➢ *Bayesian network*
  o Definition

A BN represents a set of variables in the form of nodes on a Directed Acyclic Graph (DAG). It maps the conditional independencies of these variables.
The BN is defined by:
  ❖ A DAG G= (V, E), where V is a set of nodes of G, and E is a set of arcs of G.
  ❖ A finite probabilistic space $(\Omega, Z, p)$.
  ❖ A set of random variables associated with graph nodes and defined on $(\Omega, Z, p)$ as :

$$p(V_1, V_2, ...., V_N) = \prod_{i=1}^{N} p(V_i \mid C(V_i)) \quad (1)$$

where $C(V_i)$ is a set cause [parents] $V_i$ in the graph G.

➢ *Hierarchical Bayesian Network*

The HBNs are a generalization of the standard BNs, where a node in the network may be an aggregate data type. This allows the random variables of the network to arbitrarily represent structured types. Within a single node, there may also be links between components, representing probabilistic dependencies among parts of the structure. The HBNs encode the conditional probability dependencies in the same way as the standard BNs.

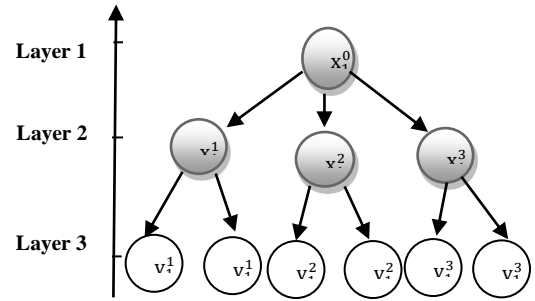

Fig. 4   Hierarchical Bayesian network with three layers

➢ *Dynamic hierarchical Bayesian network*

The DHBNs are an extension of the static HBNs which represent the temporal evolution of any random variable. Thus, a dynamic BN is a chain of the same BN repeated as many times as needed. The temporal dynamics are represented by arcs connecting the various static BNs between each other. The construction of a DHBNs requires the determining its structure and its parameters. So, to specify a DBN [6], we need to define the intra-slice topology (Within a slice), the inter-slice topology (between two slices), as well as the parameters for the first two slices as follows:

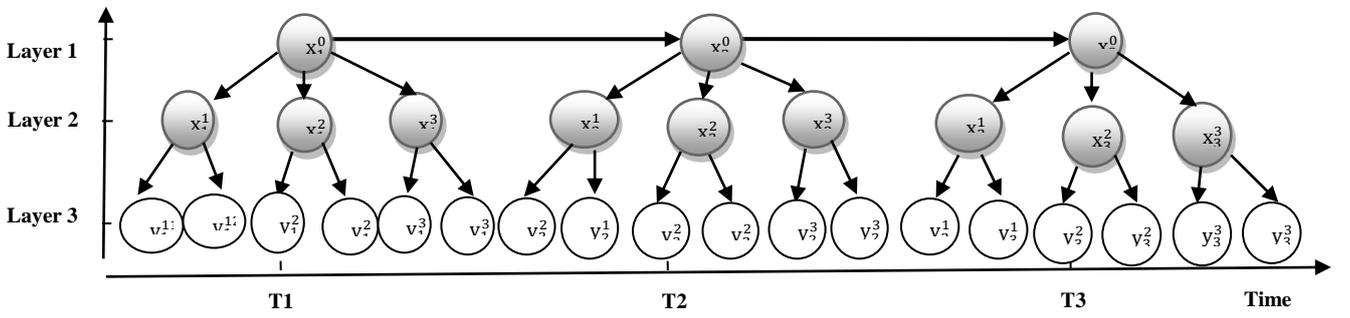

Fig. 5 Dynamic hierarchical Bayesian network for a word image composed of three characteres

### Inference and learning in dynamic Bayesian network

This particular DBN is equivalent to a traditional HMM because it emulates what an HMM does. One major difference, is that it explicitly represents the HBN consistence of the word, character, and sub character.

❖ Structural learning : Description of settings

- $X_t = \{x_t, x_t^1, x_t^2, x_t^3 \ldots, x_t^N\}$ is a set of the hidden state in the time t, where N is the number of the latent states in one t slice.

- $Y_t = \{\{y_t^{11}, y_t^{12}, \ldots, y_t^{M^1}\}, \{y_t^{21}, y_t^{22}, \ldots, y_t^{2M^2}\}, \ldots, \{y_t^{M^N 1}, \ldots, y_t^{M^N N}\}\}$ is a set of the observed state in the time t, where each state has two observed states, with $M^j \; \forall \, j \in [1, N]$ being the number of the observed state of each latent sate. We can assume that the set of $Y_t = \{Y_t^1, Y_t^2 \ldots Y_t^N\}$

- $\pi = \{\pi_i\}$ the initial state probability

$$\pi_i = P(X_1 = i) = \prod_{j=1}^{N-1} P(x_1^j = j | x_1 = i), 1 < i < N$$

- $A = \{a_{ij}\}$: the state transition probability:
$$a_{ij} = P(x_t = j | x_{t-1} = i), 1 < i, j < T$$

- $B = \{b_j(k)\}$ : the observation symbol probability distributions:
$$b_j(k) = P(Y_t = k | X_t = j) = \prod_{i=1}^{N} P(Y_t^i = k | X_t = j)$$

❖ Parameter Learning

More succinctly, a HDBN can be represented by the parameter $\lambda = (A, B, \pi)$. To suitably use the HDBN in the handwriting recognition, three problems must be solved. The first is concerned with the probability evaluation of an observation sequence, given the model $\lambda$. In the second problem, we attempt to determine the state sequence that best explains the input sequence of observations. The third problem consists in determining the method to optimize the model parameters to satisfy a specific optimization criterion. The model parameter determination is usually done by the Expectation/Maximization procedure, and consists in iteratively maximizing the observation, given the model, and often converges to a local maximum. As the DBN usually captures the joint distribution of the variable sequence, it is typically learned by maximizing the log likelihood of the training sequence $\theta_{MLE} = \arg \max_{\theta} P(Y|\theta)$

❖ Inference and Recognition :

The simplest inference method for a discrete-state DBN is to convert it to an HMM, and then to apply the forwards-backwards algorithm. Therefore, the probability $P(O|\lambda)$ of a DBN models with an explicit state duration, for an observation vector sequence using the length of the observation sequence for each t slice, can be computed by a generalized forward-backward algorithm. This choice is justified by the availability of the estimation formulas, which are derived with respect to the likelihood criterion, for the parameter set of the distribution. In the recognition phase, a solution to the state decoding problem, based on dynamic programming, has been designed, namely Viterbi algorithm [15]. The sequence of the extracted feature vectors is passed to a network of lexicon formed of word models. This algorithm has time complexity $O(T.N^2)$ and taking only $O(N^2.\log T)$ memory (where T is the length of sequence and N is the number of symbols in the state).

## IV. EXPERIMENTATION AND DISCUSSION

In this section, we will present the experimental results. The tests have been done on a corpus of Tunisian city names that is extracted from the IFN/ENIT database. We have selected a subset of 14 models and have generated a database of 5600 words, composed of 400 different images per model. Each city name word corresponds to a model. On this database, we have defined various training and test sets of different sizes. We have used the samples of the sub basis "a" and "b" for the training, those of the sub basis "c" for the validation, and those of the sub basis "d" for the test.

After pre-processing, the optimal width of the smoothed histogram is determined empirically. We have chosen the width that provides a number of characters, which reflects better the number of characters in the model.

After segmenting the word into characters, we divide the character block into cells. The optimal cell number is determined empirically using different numbers varying from 2 to 8. The rate recognition is listed in the figure bellow. We have chosen the cell number that provides the highest recognition rate.

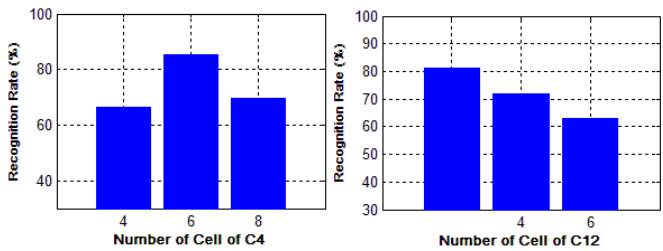

Fig. 6 Rate recognition with different number of cell of C4 'نقة' and C12 'شماخ'

After that, an optimal number of states used in the DBN is also determined empirically. Using possible numbers varying equally from 9 to 25, the obtained recognition rates are listed in Fig.7. It has been noted that the recognition rate improves as the number of the states increases to reach the maximum possible state for a specific feature set. This makes the training data independent from the testing data, hence avoiding the over-fitting the classifier to test the data. In our case, as shown in the figure below, the optimal state number of class c3 is found as 21, and of the class c4 is found as 13.

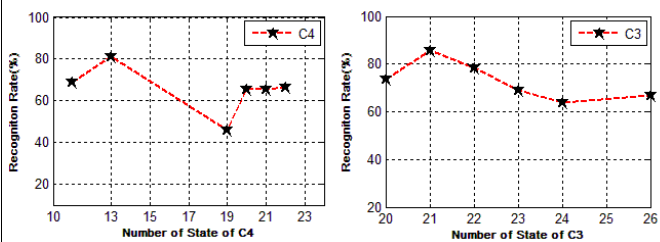

Fig. 7 Recognition rate with different number of states for two classes

In this paper, a DBN has been used for the Arabic handwritten word. Each character is represented by its feature vector, and each character requires a number of observations for training and testing the DBN. In the phase of quantizing the data, the experiments have been conducted using an 8 codebook size parameter value: 6,18,24,36,48,58,68,100. Fig.8 shows the result for a different codebook size that yields to a better recognition rate of a 'شماخ' and 'الرضاع' class. The best performance has been found using the number of observation sizes which is 58 for C4 and 38 for C7.

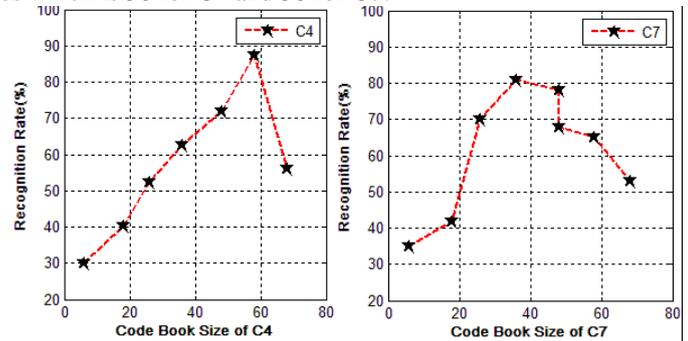

Fig. 8 Recognition rate versus codebook size.

Fig.9 shows the experimental results of the performance evaluation of our recognition system using the training and test sets. This leads to an average recognition rate of about 91.25%, which is achieved with the training set and of about 81.5% which is achieved with the test set.

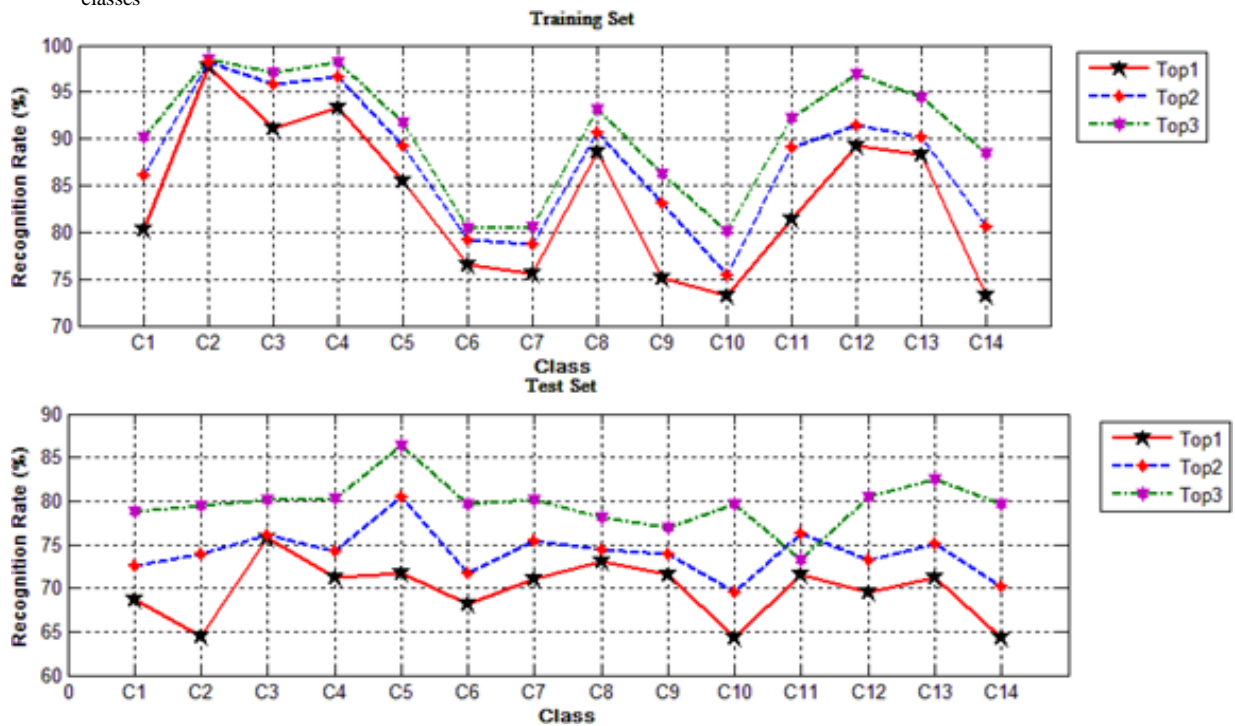

Fig. 9: Recognition rates obtained by Training and test sets for some classes

In the figure bellow, we show the precision and recall curve.

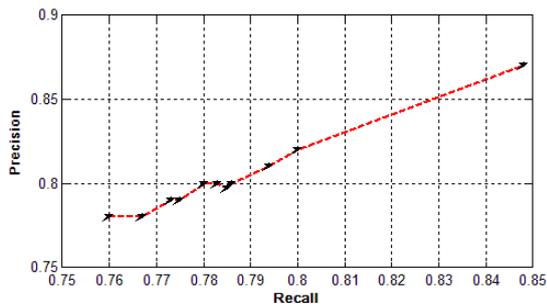

Fig.10: Precision/recall curve

The database is separated into four sets, a, b, c, and d, in order to perform four cross validation experiments.

TABLE I  TESTING WITH CROSS VALIDATION

| Learning Set | Test Set | Recognition Rate |
|---|---|---|
| b, c, d | a | 80.9% |
| a, c, d | b | 81.23% |
| a, b, d | c | 80.46% |
| a, b, c | d | 82 % |
| Recognition Rate | | 81.14% |

➢ Error analysis and comparaison

A preliminary error analysis shows different causes for the classification errors according to the nature of the Arabic cursive handwriting words. Besides, the most likely error is due to the segmentation step because the descenders of a letter and the diacritical marks are not often at the exact position on top or under the main letter. In addition, each word has a number of characters which has the same length of the inter-slice, this contributing to the recognition rate error. Also, some character shapes are insufficiently represented in the database, as results their models are poorly trained. Accordingly, Table II illustrates a comparative study.

TABLE II  COMPARATIVE RESULTS USING IFN/ENIT DATABASE

| Authors | Classifier | Lexicon usage | Word recognition rate (%) | |
|---|---|---|---|---|
| | | | Single classifier | Multi classifiers |
| Margner et al. | HMM | yes | 74.69 | |
| hamdani et al. | Multiple HMM | Not mentioned | 49-63 | 81.93 |
| Kessentini et al. | Multi-stream HMM | yes | 63.5-70.5 | 79.6 |
| Parvez et al. | FATF with set-medians | yes | 79.58 | |
| Proposed | DHBN | yes | 80.4-82 | |
| Giménez et al. | windowed Bernoulli HMMs | Not mentioned | 84 | |

V. CONCLUSION

We have proposed a new approach for the offline Arabic handwritten word recognition based on the dynamic hierarchical Bayesian network using a free segmentation released by a smoothed vertical projection histogram with different width values. The model is composed of three levels. The first level represents the layer of the hidden node which models the character class. The second layer models a frame set representing the sub-characters, and the third layer models the observation nodes. The developed system has been experimented and the results are provided on a subset of the benchmark IFN/ENIT database. These results show a significant improvement in the recognition rate due to the use of the dynamic hierarchical Bayesian networks. Most of the recognition errors of the proposed system can be attributed to the segmentation process error and to the poor quality of some data samples. Also, some characters shapes are insufficiently represented in the database. Consequently, their models are poorly trained.